\documentclass[letterpaper, 10 pt, conference]{ieeeconf}  

\IEEEoverridecommandlockouts                              

\usepackage{graphics} 
\usepackage{epsfig} 
\usepackage{mathptmx} 
\usepackage{amsmath} 
\usepackage{amssymb}  
\usepackage{lipsum}
\usepackage{multicol}
\usepackage{multirow}
\usepackage{booktabs}
\usepackage{stfloats}
\usepackage{float}
\usepackage{color,xcolor}
\usepackage{geometry}
\usepackage{hyperref}
\usepackage{ulem}

\begin{document}
\title{\LARGE \bf
RL-GSBridge: 3D Gaussian Splatting Based\\ Real2Sim2Real Method for Robotic Manipulation Learning
}


\author{Yuxuan Wu$^{*}$, Lei Pan$^{*}$, Wenhua Wu, Guangming Wang,  Yanzi Miao, Fan Xu$^{\#}$ and Hesheng Wang$^{\#}$ 
\thanks{This work was supported in part by the Shenzhen Science and Technology Program under Grant KJZD20230923114812027. (Corresponding Author: Fan Xu and Hesheng Wang)}
\thanks{* Equal contributions}
\thanks{\# Co-corresponding authors}
\thanks{
Y. Wu, W. Wu, F. Xu, and H. Wang are with the Shenzhen Research Institute of Shanghai Jiao Tong University, Shenzhen 518000, China. }
\thanks{Y. Wu, F. Xu, and H. Wang are also with the Department of Automation, Shanghai Jiao Tong University, Shanghai 200240, China. (e-mail: furrygreen@sjtu.edu.cn; xufanlyra@sjtu.edu.cn; wanghesheng@sjtu.edu.cn) W. Wu is also with MoE Key Lab of Artificial Intelligence, AI Institute, Shanghai Jiao Tong University, Shanghai 200240, China.}
\thanks{L. Pan and Y. Miao are with the School of Information and Control Engineering, China University of Mining and Technology, Xuzhou 221100, China. }
\thanks{G. Wang is with the Department of Engineering, University of Cambridge, Cambridge CB2 1PZ, U.K. (e-mail: gw462@cam.ac.uk)}%
\thanks{Code is available at 
\href{https://github.com/IRMV-Manipulation-Group/RL-GSBridge}{https://github.com/IRMV-Manipulation-Group/RL-GSBridge}}
%
}

\maketitle
\begin{abstract}
Sim-to-Real refers to the process of transferring policies learned in simulation to the real world, which is crucial for achieving practical robotics applications. However, recent Sim2real methods either rely on a large amount of augmented data or large learning models, which is inefficient for specific tasks. In recent years, with the emergence of radiance field reconstruction methods, especially 3D Gaussian splatting, it has become possible to construct realistic real-world scenes.
To this end, we propose RL-GSBridge, a novel real-to-sim-to-real framework which incorporates 3D Gaussian Splatting into the conventional RL simulation pipeline, enabling zero-shot sim-to-real transfer for vision-based deep reinforcement learning. We introduce a mesh-based 3D GS method with soft binding constraints, enhancing the rendering quality of mesh models. Then utilizing a GS editing approach to synchronize the rendering with the physics simulator, RL-GSBridge could reflect the visual interactions of the physical robot accurately. 
Through a series of sim-to-real experiments, including grasping and pick-and-place tasks, we demonstrate that RL-GSBridge maintains a satisfactory success rate in real-world task completion during sim-to-real transfer.
Furthermore, a series of rendering metrics and visualization results indicate that our proposed mesh-based 3D GS reduces artifacts in unstructured objects, demonstrating more realistic rendering performance.


\end{abstract}


\section{INTRODUCTION}
Learning robotic action policies in simulation and transferring them to real-world represents an ideal robotic learning strategy that balances both the cost and safety of the learning process.
However, a significant bottleneck is the reliability of sim-to-real transfer, which impacts the potential of the entire framework towards substantial challenges.

With the continuous development of the simulation-to-reality(Sim2Real) field, extensive work is advancing progress from multiple perspectives \cite{zhao2020towards, muratore2020bayesian, arndt2020meta, traore2019continual}. However, most Sim2Real methods attempt to expand the distribution of training data to cover various situations that may arise in reality, or to train a highly generalized large model to learn knowledge for different tasks. This significantly increases the difficulty in training stage.

\begin{figure}[t]
    \centering
    \includegraphics[width=1\linewidth]{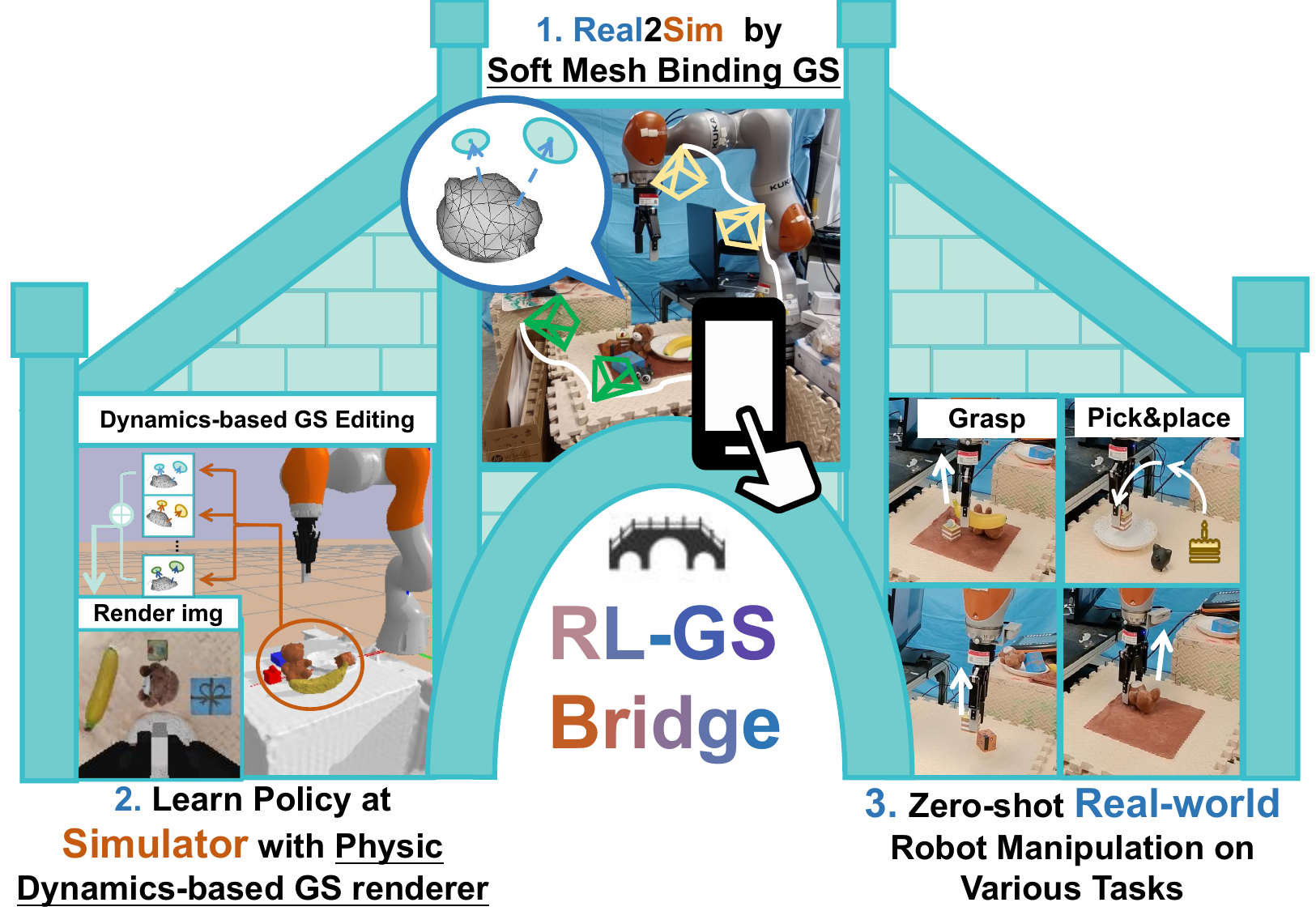}
    \caption{Pipeline of RL-GSBridge. (1) Real2Sim Environment Transfer. Real-world scenarios is reconstructed through a novel soft mesh binding GS model. (2) Learn Policy at Simulator with GS Render. With physical dynamics-based GS editing, RL policies learn through realistic rendered images in simulation. (3) Zero-shot Real-world Robot Manipulation. We directly apply the policy to real-world tasks without fine-tuning.}
    \label{fig:show}
\end{figure}

To avoid additional training burden and achieve ideal Sim2Real performance on specific tasks, a novel framework is needed.
Recently, advances in radiance field-based reconstruction methods \cite{mildenhall2020nerf, zhang2020nerf++, zhu2022nice, zhu2023sni} provide new directions for Sim2Real training. Based on a simple idea—using radiance field reconstruction to create a visually realistic robot training environment—can we achieve satisfactory Sim2Real performance? For this purpose, we design a Real2Sim2Real visual reinforcement learning framework, RL-GSBridge, that bridges the real-to-sim gap by 3D Gaussian splatting, as shown in Fig. \ref{fig:show}. 
Utilizing 3D Gaussian splatting and editing techniques, RL-GSBridge provide a `virtual-reality' simulation platform for policy learning.


For vision-based robot tasks, it is crucial to avoid illusions caused by inconsistencies between visual perception and contact geometry. Thus, it is required to ensure an accurate geometric representation of the model while achieving more realistic rendering results.
GaMeS \cite{waczynska2024games} has drawn our attention as it is a mesh-based Gaussian splatting (GS) method, which ensures that the optimization of GS units is performed within the geometric mesh model.  
However, it enforces Gaussians to be aligned with the mesh grid planes, which could be considered as ‘hard mesh binding', thereby limiting the flexibility of GS units.
To address this, we propose a soft mesh binding method for Gaussian Splatting, which could further enhance rendering quality while preserving editing capabilities for both objects and the background.

Physics-based dynamics simulation is also a crucial and challenging aspect of sim2real. To tackle this, an off-the-shelf physics simulator is used to provide dynamic changing information. The GS editing process simultaneously updates the scene, ensuring that the rendering results align with physical interaction processes.


Based on the designed simulation platform, we train robotic manipulation policies by deep reinforcement learning methods. The model is trained on grasping and pick-and-place tasks across scenarios that include diverse textures, geometric shapes, and patterned desktop backgrounds.

Under the RL-GSBridge framework, the policy shows a minor variation in success rates during Sim2Real transfer. This means that the policy maintains effective performance on real-world tasks, reflecting a strong ability to generalize from simulation to real-world environments.

To summarize our contributions:
\begin{itemize} 
\item \textbf{A Novel Sim2Real RL Framework}: Leveraging the high-fidelity rendering of 3D GS and the convenience of modeling scenes with only consumer-grade cameras.

\item \textbf{A Soft Mesh Binding GS Modeling Method}: Proposing a soft mesh binding strategy to replace the hard mesh binding baseline, enhancing the flexibility and render quality.

\item \textbf{Physical Dynamics-Based GS Editing}: Integrating dynamics signals from the simulator to edit 3D GS models, reflecting realistic physical robotic interactions.

\item \textbf{Validation on Real Physical Robots}: Testing the RL-GSBridge framework on physical robots through grasping and pick-and-place tasks in real-world scenarios with complex textures and geometries. 
\end{itemize}




\section{RELATED WORK}
\subsection{Sim2Real Transfer in RL}
RL constructs an interactive learning model in which an agent learns to maximize rewards through trial and error. By incorporating deep learning, deep reinforcement learning(DRL) enhances the framework's ability to tackle more complex tasks \cite{franccois2018introduction}. DRL has shown impressive performance across various domains, including games \cite{silver2016mastering}, finance \cite{deng2016deep}, autonomous driving \cite{pan2017virtual}, and robotics \cite{pinto2017asymmetric}. However, most applications are restricted to virtual environments due to real-world constraints related to safety, efficiency, and cost.

To improve the feasibility of deploying models in the real world, many researchers strive to bridge the gap between simulation and reality\cite{liu2024aligning}. These methods include domain randomization \cite{tobin2017domain} and domain adaptation \cite{bousmalis2018using, fang2024your}. Domain randomization involves varying task-related parameters in the simulation to cover a broad range of real-world conditions, while domain adaptation focuses on extracting a unified feature space from both sources. Higher-level learning methods include metalearning \cite{wang2016learning} and distillation learning \cite{rusu2015policy}. Meta-learning aims to teach robots the ability to learn new tasks, whereas distillation learning trains a student network using knowledge from expert networks.

Our approach aligns with the first category of gap-bridging methods, but with a unique twist. We use soft mesh binding GS to create realistic simulation environments for robot training. 
Typically, achieving such fidelity requires expensive 3D scanning equipment or CAD expertise. In contrast, GS models visually realistic simulation environments using only multi-view images captured with consumer-grade devices.

\subsection{Radiance Field in Robotics}
Neural Radiance Fields (NeRF)~\cite{mildenhall2020nerf} is an implicit representation technique for novel view synthesis. It optimizes the parameters through multi-view images, and allows for the synthesis of any target views using volumetric rendering. 
NeRF's representation has been applied to various tasks, including Simultaneous Localization and Mapping (SLAM)~\cite{zhu2022nice, zhu2024sni, zhu2024semgauss}, scene reconstruction~\cite{yu2022monosdf}, scene segmentation~\cite{zhi2021place}, navigation ~\cite{adamkiewicz2022vision}, and manipulation~\cite{dai2023graspnerf}.

NeRF-RL~\cite{driess2022reinforcement} treats the novel view synthesis task of NeRF as a proxy task, where the learned encoding network is directly used as feature input for reinforcement learning. Y. Li \textit{et al.} \cite{li20223d} uses NeRF as a perceptual decoder for the hidden states of a world model, training an additional dynamics estimation network to predict future state changes. NeRF2Real \cite{byravan2023nerf2real} learns real-world contexts by converting background meshes into a simulator, and trains robots to perform visual navigation, obstacle avoidance, and ball-handling tasks.
However, the training and rendering speed of Vanilla NeRF has consistently been a bottleneck limiting its further deployment in practical applications.

3D GS \cite{3dgs} is an explicit radiance field method that directly updates the attributes of each 3D Gaussian component to optimize scene representation. 3D GS employs a splatting technique.
for rendering, and achieve extremely fast training efficiency through CUDA parallel technology. Moreover, compared to the implicit representation of NeRF, the explicit representation facilitates tracking dynamic scene modeling and editing scene content.

Many robotics works also incorporate 3D Gaussian techniques for perception and motion learning~\cite{zhu20243d}. ManiGaussian~\cite{lu2024manigaussian} uses 3D GS as visual and dynamic scene representation for policy learning. 
Quach \textit{et al.} \cite{quach2024gaussian} combine GS with Liquid networks for real-world drone flight navigation tasks, training in simulation and then deploying the policy in the real world. 


In contrast to NeRF2Real \cite{byravan2023nerf2real}, which directly textures the foreground objects in simulator, we model each foreground object with editable GS parameters. Also unlike the method designed by Quach \textit{et al.} \cite{quach2024gaussian}, which trains drone navigation policies for target-oriented tasks, our approach involves robotic arm manipulation tasks with complex interactions.




\section{METHODS}


RL-GSBridge aims to harness the potential of high-fidelity 3D GS models in Sim2Real for robot action training tasks. 
In this paper, we focus on manipulation tasks for robotic arms. 
As shown in Fig. \ref{fig:show}, the overall framework is divided into two parts: Real2Sim and Sim2Real. 
In Real2Sim stage, we collect real-world image data $\{ {\mathbf{I}_k}\}$, where $\mathbf{I}_k$ represent the image sequences of the $k$-th object or background in the scenarios. We will model both the geometry and appearance of the scene to build a simulation environment for the manipulation task. 
In Sim2Real stage, we use visual perception and deep reinforcement learning to train a policy network, and directly transfer the policy to the real world.

\subsection{Real2Sim: Building simulator with soft mesh binding GS}
To obtain a realistic simulation environment, we use consumer-grade cameras to capture 2D image data $ \mathbf{I}_k $ of desktop-level operating platforms. Our goal is to model a geometrically accurate and texture-realistic simulation model for training operational tasks. More specifically, we use mesh models $\{ {\mathbf{M}_k}\}$ to represent the accurate geometric information, and Gaussian sets $\mathbf{G}_k = \{ \mathbf{g}_k^i\}$ for high-quality texture.
Below, we sequentially describe the steps to construct the Simulator with GS renderer.

\subsubsection{Real-world Data Preparing}

We use a monocular camera or mobile phone to capture a 1-2 minute video of the target object on the experimental platform. 
We select approximately 200 keyframes from the video and use 
COLMAP \cite{schoenberger2016sfm} to obtain the camera's internal and external parameters for each frame. Image segmentation algorithm is also needed for extracting the target object, and we use an off-the-shelf segmenter, SAM-track \cite{cheng2023segment}, for efficient object segmentation.

To complete the GS model reconstruction for the simulator, it is also necessary to pre-model a geometrically accurate mesh model as a prior model for GS. 
Here we consider a classic and stable open-source package openMVS to obtain the corresponding mesh model. 

\subsubsection{Real2Sim modeling by soft mesh binding GS}
With the object mesh, we can define Gaussian units within triangular faces as in GaMeS \cite{waczynska2024games}, a method that binds Gaussians onto the surface of meshes, and optimizes their properties through multi-view consistency. Vanilla GS \cite{3dgs} optimizes the attribute parameters of Gaussian units located at each point cloud position, where $\theta^i = (\mu^i, r^i, s^i, \sigma^i, c^i)$ denotes the position, rotation, scale, opacity, and color of the $i$-th Gaussian unit $ \mathbf{g}^i $, respectively. To achieve controllable geometric editing effects, GaMeS \cite{waczynska2024games} constrains Gaussian units within the triangular mesh of the object surface, using the mean vector as the convex combination of the mesh vertices to establish the positional relationship between the Gaussian unit and the three vertices of the triangular mesh:

\begin{equation}
\begin{aligned}
\mu^i\left( {{\alpha _1^{i}},{\alpha _2^{i}},{\alpha _3^{i}}} \right) = {\alpha _1^{i}}{{\rm{\textbf{v}}}_1} + {\alpha _2^{i}}{{\rm{\textbf{v}}}_2} + {\alpha _3^{i}}{{\rm{\textbf{v}}}_3},
\end{aligned}
\end{equation}
\begin{figure}[t]
    \centering
    \includegraphics[width=0.8\linewidth]{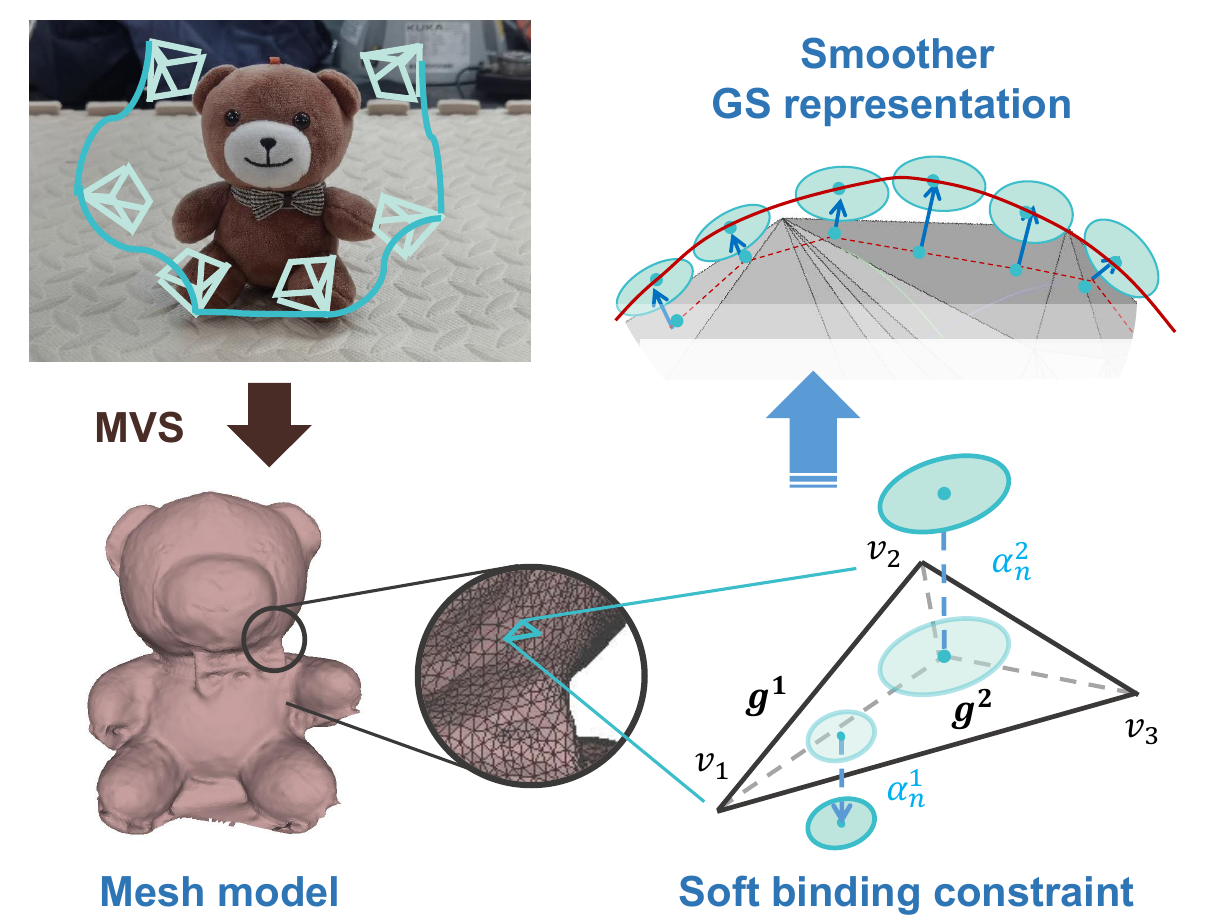}
    \caption{Mesh-based GS Reconstruction with Soft Binding Constraints: Releasing the hard constraints of GaMeS \cite{waczynska2024games} in the normal direction for smoother and more flexible object surfaces.}
    \label{fig:somebigs}
\end{figure}
here, $\mathbf{v}_1, \mathbf{v}_2, \mathbf{v}_3$ represents the triangular mesh vertex positions, $\alpha_1^{i}, \alpha_2^{i}, \alpha_3^{i}$ are learnable positional weight parameters for $ \mathbf{g}_i $. However, this approach of enforced mesh-GS binding would diminish the flexibility of 3D Gaussians, limiting the optimization of Gaussian units when the mesh model is not accurate, introducing certain undesirable defects.

To address this, we propose a soft mesh binding method, which builds upon GaMeS \cite{waczynska2024games}, but relaxes the enforced hard mesh binding to a soft binding constraint. We introduce a component along the normal direction into the vector of positions, specifically:
\begin{equation}
\begin{aligned}
\mu^i\left( {{\alpha _1^{i}},{\alpha _2^{i}},{\alpha _3^{i}}, {\alpha _n^{i}}} \right) = {\alpha _1^{i}}{{\rm{\textbf{v}}}_1} + {\alpha _2^{i}}{{\rm{\textbf{v}}}_2} + {\alpha _3^{i}}{{\rm{\textbf{v}}}_3} + {\alpha _n^{i}}{{\rm{\textbf{v}}}_n},
\end{aligned}\label{eq2}
\end{equation}
here, $\alpha _n^{i}$ is a learnable weight parameter and $\rm{\textbf{v}}_n$ is the normal vector. $\alpha _n^{i}$ is constrained within the range of $[-1, 1]$, ensuring the association between each mesh model element and Gaussian pairs. As shown in Fig. \ref{fig:somebigs}, our method allows the Gaussian units within the mesh to float within a certain range along the normal vector. This flexibility in Gaussian unit optimization could
bring a smoother distribution of Gaussian units on the object's surface. Ultimately, the algorithm not only ensures that the Gaussian model can still represent the accurate geometric structure according to the mesh models, but also offers some tolerance and refinement space. Besides, the binds between the mesh grid and Gaussians could even provide the possibility to handle non-rigid objects, as demonstrated in section \ref{sec_soft}.
\subsubsection{Physic Dynamics-Based GS Editing}
After obtaining the visual GS models $\{\mathbf{G}_k\}$ and geometry models $\{\mathbf{M}_k\}$ for real-world objects, we combine the dynamic simulation results from the simulator with real-time Gaussian model updates and render views to ensure that the visual representation follows the entire physical change process. We first use RANSAC plane regression and manual alignment methods to align 
GS models with mesh models in the simulator, and set the initial position of the GS model.


With the aligned initial model, we read the real-time pose changes of each object in the operational scene from the simulator. For the $k$-th object with its GS model $\{\mathbf{G}_k\}$, We acquire the rotation quaternion $ q_k $  and the homogeneous transformation matric $\mathbf{T}_k $ in the world coordinate system. Given Gaussian parameters $\theta_k^i = (\mu_k^i, r_k^i, s_k^i, \sigma_k^i, c_k^i)$, The editing process of each Gaussian $\mathbf{g}_k^i$ in the model set $\mathbf{G}_k$ could be formulated as:
\begin{equation}
\begin{aligned}
    &\mu_k^{i} = \mathbf{T}_k \mu_k^{i},&r_k^{i} = q_k \times r_k^{i}.
\end{aligned}
\label{sim trans}
\end{equation}
As for local non-rigid transformations on the mesh grid, the Gaussians could be updated according to Equation \ref{eq2}. After applying transformations to all object models, we concat the Gaussian sets to acquire the Gaussian model of the whole scene $\mathbf{G}_{scene}$. Then we use rasterization to render $\mathbf{G}_{scene}$, obtaining the synchronized edited rendering view. 

\begin{figure*}
    \centering
    \includegraphics[width=0.75\linewidth]{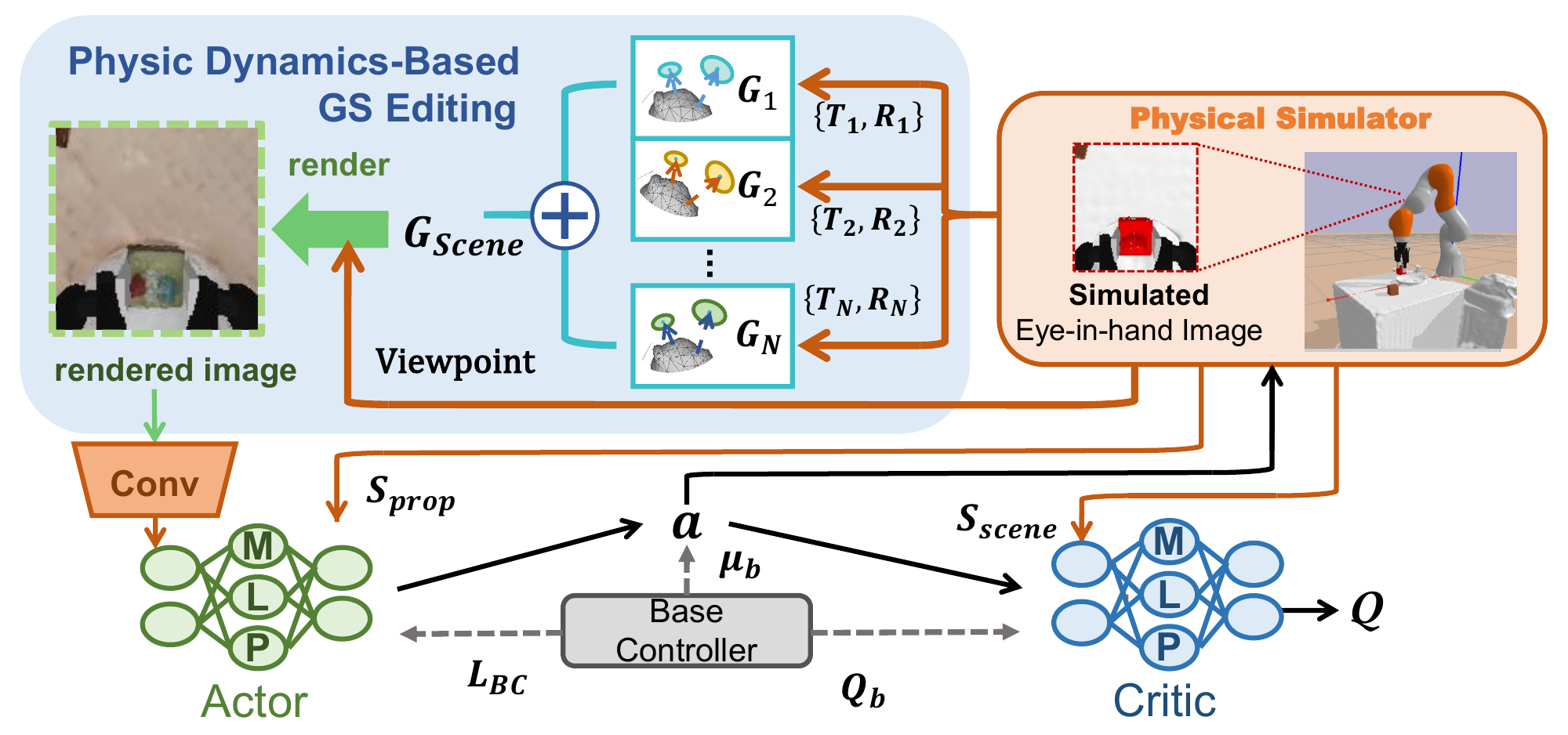}
    \caption{Policy training pipeline in RL-GSBridge. In the upper half of the figure, physic dynamics-based GS editing receives the transformation signals of objects and synchronizes the states of GS models. In the lower half of the figure, an actor-critic RL network receives first-person perspective images rendered by GS models as input, to learn a vision-based manipulation policy.}
    \label{fig:pipeline}
\end{figure*}

\begin{table}[!t]
\centering
\caption{Comparison of Success Rates Between RL-GSBridge and RL-Sim in Grasping Experiments, All Conducting Under Foam Pad(FP) Background. \textbf{Bold} Indicates Better Results. Values in Parentheses Represent Relative Change in Success Rate During Sim2Real Transfer. \textcolor[HTML]{32CB00} {(↓xx\%)} Indicates a Decrease, While \textcolor[HTML]{FE0000} {(↑xx\%)} Indicates an Increase.}
\label{tab:RLcontrast}
\small
\renewcommand{\arraystretch}{1.3}
\setlength{\tabcolsep}{5pt}
\resizebox{1\columnwidth}{!}{%
\begin{tabular}{lccccc}
\toprule
\multicolumn{1}{l}{Object} & \multicolumn{2}{c}{Small\_cube} & \multicolumn{2}{c}{Bear} \\ \cline{2-5}
\multicolumn{1}{l}{Test scene} & Sim & Real & Sim & Real \\
\midrule
RL-sim & \textbf{96.88} & 12.50 \color[HTML]{32CB00} (↓87\%)& \textbf{93.75} & 25.00 \color[HTML]{32CB00} (↓73\%) \\
RL-GSBridge& \textbf{96.88} & \textbf{96.88} & 87.50 & \textbf{100.00} \color[HTML]{FE0000} (↑14\%)\\

\bottomrule
\end{tabular}
}
\end{table}

\begin{table*}[t]
\caption{Sim2Real Results for Grasping Task in Various Manipulation Scenarios. Contents in Parentheses After the Object Names Represent Different Backgrounds(Bg), Where FP Denotes Foam Pad, and TC Denotes Tablecloth.}
    \begin{center}
    \label{tab:testgrasp}
\centering
\normalsize
\renewcommand{\arraystretch}{1.7}
\setlength{\tabcolsep}{2.2pt}
\resizebox{2\columnwidth}{!}{%
\begin{tabular}{lcccccccccccc}
\toprule
Object (Bg) & \multicolumn{2}{c}{Cake (FP)} & \multicolumn{2}{c}{Banana (FP)} & \multicolumn{2}{c}{Small\_cube (TC)} & \multicolumn{2}{c}{Cake (TC)} & \multicolumn{2}{c}{Banana (TC)} & \multicolumn{2}{c}{Bear (TC)} \\
\cline{2-13} 
Test scene & \multicolumn{1}{c}{Sim} & Real & \multicolumn{1}{c}{Sim} & Real & \multicolumn{1}{c}{Sim} & Real & \multicolumn{1}{c}{Sim} & Real & \multicolumn{1}{c}{Sim} & Real & \multicolumn{1}{c}{Sim} & Real \\
\midrule
Success rate (\%) & \multicolumn{1}{c}{\textbf{100.00}} & \textbf{100.00} & \multicolumn{1}{c}{\textbf{100.00}}& 93.75 \color[HTML]{32CB00} (↓6\%)& \textbf{96.88}  & 87.50 \color[HTML]{32CB00} (↓10\%)& \multicolumn{1}{c}{\textbf{100.00}}& 93.75 \color[HTML]{32CB00} (↓6\%)& \multicolumn{1}{c}{\textbf{100.00}} & 96.88 \color[HTML]{32CB00} (↓3\%)& \multicolumn{1}{c}{\textbf{87.50}} & 75.00 \color[HTML]{32CB00} (↓14\%)\\
\bottomrule
\end{tabular}%
}
    \end{center}
\end{table*}

\begin{table}[hbt]
\centering
\caption{The Sim2Real Result for the Pick and Place Task.}
\label{tab:testpickplace}
\small
\renewcommand{\arraystretch}{1.5}
\setlength{\tabcolsep}{25pt}
\resizebox{1\columnwidth}{!}{%
\begin{tabular}{lcc}
\toprule
Object & \multicolumn{2}{c}{Cake \& Plate} \\ \cline{2-3} 
Test scene & \multicolumn{1}{c}{Sim} & Real \\
\midrule
\multicolumn{1}{c}{Success rate (\%)} & 68.75 & 71.87 
 {\color[HTML]{FE0000} (↑4\%)} \\
\bottomrule
\end{tabular}%
}
\end{table}

\subsection{Sim2Real: Train in simulation with physic dynamics-based GS renderer and zero-shot transfer to reality}
We use Pybullet \cite{coumans2016pybullet} as the simulation training platform, and employ SAC (Soft Actor-Critic) \cite{haarnoja2018soft} algorithm for policy learning, due to its mature development in RL and widespread application in robotics.
The SAC is an offline policy algorithm belonging to maximum entropy RL, which consists of two critic networks, $Q_{\phi, 1}(s, a)$ and $Q_{\phi, 2}(s, a)$, and one actor network, $\pi_{\theta}(s)$. For the sampled set $B$ from the replay buffer, the update loss of critic is defined as:
\begin{equation}{ Loss_{critic} } = \frac{1}{{\left| B \right|}}\sum {_{({s_{t,}}{a_{t,}}{r_{t + 1}},{s_{t + 1}}) \in B}} {({y_t} - {Q_{\phi ,j}}({s_{t,}}{a_t}))^2},\end{equation}
here, $({s_{t,}}, {a_{t,}}, r_{t + 1}, {s_{t + 1}})$ is a tuple belong to $B$, the $y_t$ is calculated by target $Q$ networks $Q _{\overline\phi ,j}$:
\begin{equation}y_t=r_t+\gamma(\min_{j=1,2} Q_{\overline\phi, j}(s_{t+1}, a_{t+1})-\alpha\log\pi_{\theta}(a_{t+1}|s_{t+1})).\end{equation}
In continuous action space, policy $\pi_{\theta}$ outputs actions of Gaussian distribution. 
The loss for actor is defined as:
\begin{equation}
    \begin{aligned}
{ Loss_{actor} } = \frac{1}{{\left| B \right|}}&\sum {_{({s_{t,}}{a_{t,}}{r_{t + 1}},{s_{t + 1}}) \in B}} \\ &( \alpha \log {\pi _\theta }({\widetilde a_t}\left| {{s_t}} \right.) - \min_{j=1,2} Q_{\phi, j}(s_{t}, {\widetilde a_t})),\label{actor_equ}
    \end{aligned}
\end{equation}
where ${\widetilde a_t}$ is sampled using the reparameterization trick, i.e., ${a_t} = {f_\theta }({\varepsilon _t};{s_t}),$
$\varepsilon$ is Gaussian random noise. 
In practical implementation, we set:
${f_\theta } = \tanh ({\mu _\theta }({s_t}) + {\sigma _\theta }({s_t}) \odot \varepsilon _t). $


Since vanilla SAC struggles to converge rapidly under sparse reward, to accelerate the learning process through automated guidance, we propose SACwB by referring to DDPGwB \cite{wang2022learning}, and introduce a lightly designed baseline controller to guide the policy. This approach avoids complex reward designs while eliminating ineffective action spaces.

In SACwB, the agent executes actions from the baseline controller with probability $\lambda$ and selects the best option between the baseline controller's and the actor's outputs with probability $1 - \lambda$, the objective is typically defined as: 
\begin{equation}
    \begin{aligned}
    y_t=r_t+\gamma\max
        ((\min_{j=1,2} Q_{\overline\phi, j}(s_{t+1}, a_{t+1})-\alpha\log\pi_{\theta}(a_{t+1}|s_{t+1})), \\ 
        (\min_{j=1,2} Q_{\overline\phi, j}(s_{t+1}, {\mu _b}({s_{t+1}}))-\alpha\log\pi_{\theta}({\mu _b}(s_{t+1})|s_{t+1})))
    .
    \end{aligned}
\end{equation}

For the supervision of the actor network, we introduce an additional behavior clone loss $L_{bc}$
in Equation \ref{actor_equ} to supervise the mean values of the action distribution through base controller actions, with the probability $\lambda$:
\begin{equation}
{L_{bc}} =  \left\| {{\mu _\theta }(s) - {\mu _b}(s)} \right\|^2,
\end{equation}
and with the probability $1 - \lambda$:
\begin{equation}
{L_{bc}} = \left\| {{\mu _\theta }(s) - \mu |\arg \max (\min_{j=1,2} Q_{\overline\phi, j}(s,{\mu _\theta }(s)))} \right\|^2,
\end{equation}
here, $\lambda$ is the decay factor, which gradually approaches 0 as training progresses.


The whole training pipeline in the simulator with physic
dynamics-based GS renderer is shown in Fig. \ref{fig:pipeline}. The RL algorithm provides executable actions to the physics simulator, which will return state information for actor-critic learning. The GS renderer simultaneously renders the realistic images by physic dynamics-based GS rendering, serving as observation input to the actor network. We rely solely on the first-person perspective image and proprioceptive state of the robotic arm, considering several advantages as described in \cite{hsu2021vision}. 
To enhance sim2real robustness, we add random noise to the rendered images of the GS model and randomly alter certain image attribute parameters during training for environment challenges.

After policy training in the simulator, we directly deploy the actor network onto the real robot arm, with the real-world eye-in-hand camera observations of the environment serving as visual input. The trained policy subsequently outputs the position of the end-effector and the gripper states as executable actions.

\begin{figure*}[ht]
    \centering
    \includegraphics[width=1\linewidth]{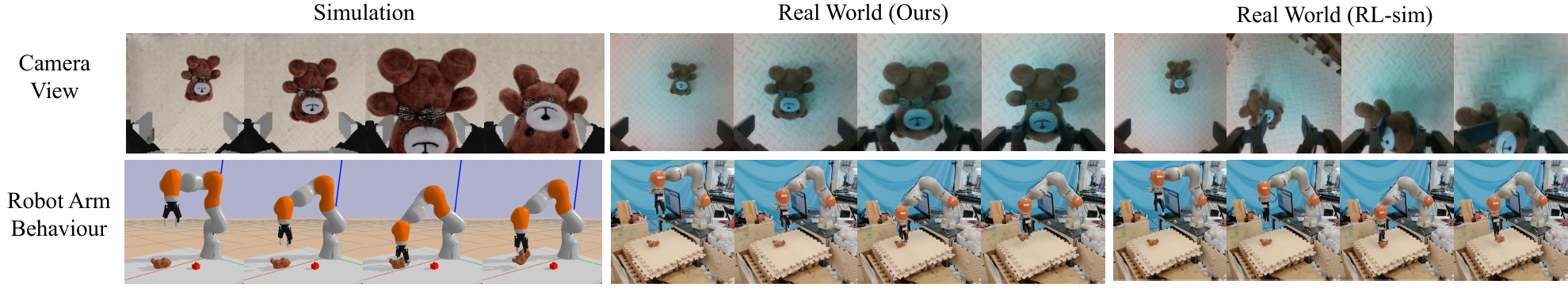}
    \caption{Comparison of sim-to-real behavior consistency between RL-GSBridge and RL-sim.}
    \label{fig:testbehaviour}
\end{figure*}

\section{EXPERIMENTS}



\begin{table*}[t]
\caption{Our Soft Mesh Binding Method Compared with GaMeS \cite{waczynska2024games} on Multiple Foreground Objects and Background Rendering Metrics. SSIM is Scale Structural Similarity Index. PSNR is Peak Signal-to-Noise Ratio. LPIPS is Learned Perceptual Image Patch Similarity. Bg Refers to the Background, While Contents in Parentheses Explain the Detailed Differences.}
\label{tab:GSresult}
\centering
\small
\renewcommand{\arraystretch}{1}
\setlength{\tabcolsep}{4.5pt}
\begin{tabular}{ccccccccccccccc}
\toprule
\multirow{2}{*}{} & \multicolumn{2}{c}{Banana} & \multicolumn{2}{c}{Bear} & \multicolumn{2}{c}{Cake} & \multicolumn{2}{c}{Small\_cube} & \multicolumn{2}{c}{Bg (Foam Pad)} & \multicolumn{2}{c}{Bg (Tablecloth)} & \multicolumn{2}{c}{Bg (Plate)} \\
 & \textbf{\textbf{Ours}} & GaMeS & \textbf{Ours} & GaMeS & \textbf{Ours} & GaMeS & \textbf{Ours} & GaMeS & \textbf{Ours} & GaMeS & \textbf{Ours} & GaMeS & \textbf{Ours} & GaMeS \\
 \midrule
SSIM↑ & \textbf{0.989} & 0.894 & \textbf{0.964} & 0.956 & \textbf{0.975} & 0.964 & \textbf{0.989} & 0.989 & \textbf{0.944} & 0.939 & \textbf{0.781} & 0.776 & \textbf{0.776} & 0.756 \\
PSNR↑ & \textbf{35.46} & 26.53 & \textbf{29.82} & 28.18 & \textbf{33.38} & 30.73 & \textbf{37.18} & 36.64 & \textbf{28.40} & 27.32 & \textbf{22.88} & 21.86 & \textbf{21.86} & 20.80 \\
LPIPS↓ & \textbf{0.026} & 0.049 & \textbf{0.034} & 0.040 & \textbf{0.066} & 0.079 & \textbf{0.012} & \textbf{0.012} & \textbf{0.144} & 0.149 & \textbf{0.248} & 0.308 & \textbf{0.308} & 0.311 \\
\bottomrule
\end{tabular}
\end{table*}

\subsection{Experiment Setup}
\subsubsection{Robot Platform}
We use a KUKA iiwa robot arm paired with a Robotiq 2F-140 gripper as the platform. The Intel RealSense D435i camera fixed on the robot arm captures the RGB images from the first-person perspective.
\subsubsection{Tasks}
As shown in Fig. \ref{fig:show}, we design two types of tasks from the robot's first-person perspective: grasping and pick-and-place operations. 

For grasping, the robot grasps a target object and lifts it up. The initial positions of the objects are
randomized within a 30 × 30 cm section. We use Small\_cube, Cake, Banana, and Bear as objects. As for the operation platform, we use a foam pad with and without a tablecloth as two different backgrounds. Success is considered when the object is lifted 10 cm above the table.

For pick-and-place tasks, the robotic arm pick up the cake model and place it on a plate. The position of the cake is set similarly as described in the grasping task. The plate is placed in a fixed position on the foam pad. Success is considered when the cake is placed on the plate.

\subsubsection{Evaluation Setup}
For each task, we conduct both the simulation and the real world experiments. During testing, we divide the 30 × 30 cm section into four quadrants. For each task, we test the policy across a fixed number of positions in each quadrant to calculate the success rate. 

\subsubsection{Baseline}
To demonstrate that our method effectively reduces Sim2Real gap, we conduct a baseline experiment called RL-sim: using the same learning method but training directly on images rendered from mesh models shaded in the PyBullet simulator. We select two representative and easily shaded objects: the Small\_cube and Bear. Grasping experiments for RL-sim are conducted on a foam pad.

\subsection{Experiment Results}
\subsubsection{Grasping}

In Table \ref{tab:RLcontrast}, we compare the grasping performance of RL strategies trained with the baseline (RL-sim) and RL-GSBridge in both simulation and real environments. For both the Small\_cube with simple geometry and textures and the Bear with complex geometry and textures, RL-sim shows a significant success rate drop when transferring to real world (an average decrease of 80\%) due to visual discrepancies. In contrast, RL-GSBridge demonstrates a minor variation in success rates, maintaining high performance as in simulation. Notably, in the Bear grasping scenario, the success rate in the real world has increased by 14.28\%. This may be attributed to the suboptimal simulation of Bear's soft material and non-structured shape in the simulation.

Table \ref{tab:testgrasp} shows that RL-GSBridge experiences an average drop of 6.6\% of the success rate in sim-to-real transfer across various complex test scenarios, including diverse objects and desktop backgrounds. This demonstrates that the integration of the physic dynamics-based GS rendering effectively bridges the perception gap between simulation and real environments, maintaining stable strategy transfer across a range of scenarios.
Additionally, we observe a noticeable decrease in success rates for the Bear in the Tablecloth background scenario, both in simulation and real environments. The primary reason for this drop is the choice of a brown tablecloth, which has similar texture features to the brown toy bear, causing difficulty in visual perception.
\begin{figure}[t]
    \centering
    \includegraphics[width=1\linewidth]{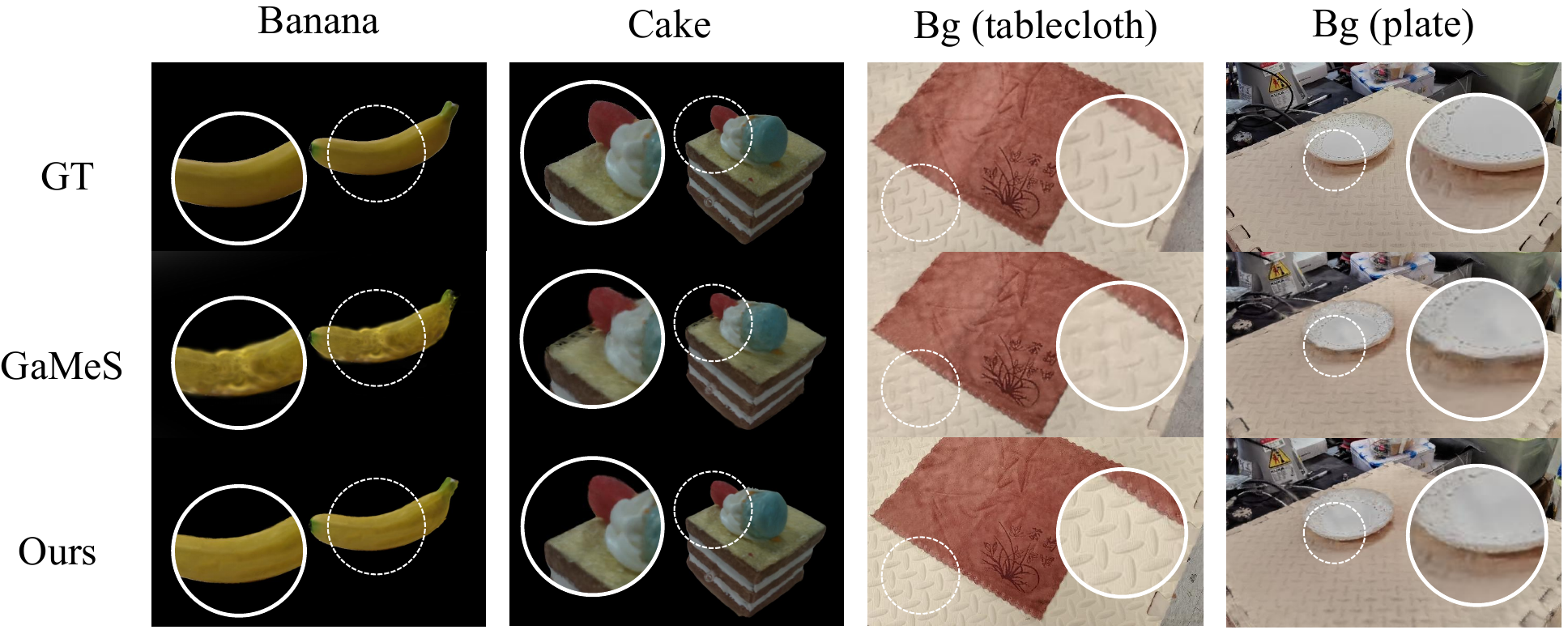}
    \caption{Our soft binding constraint reconstruction method compared with GaMeS \cite{waczynska2024games} on two foreground objects and two backgrounds.}
    \label{fig:GSresult}
\end{figure}

\subsubsection{Pick-and-Place}
As shown in Table \ref{tab:testpickplace}, we test RL-GSBridge’s Sim2Real performance in pick-and-place task where a cake is placed onto a plate. The results indicate a 4.54\% increase of success rate in real environments, mainly due to the differences in physical contacts between simulation and reality. In simulation, even minor excess contacts during the placement process are considered as task failures. In contrast, some contacts in real environments that do not affect the task can be tolerated, leading to better performance since the task is ultimately completed successfully.

\subsubsection{Comparison of Sim\&Real Behavior Consistency}
In Fig. \ref{fig:testbehaviour}, we compare the behavior of the robotic arm in the simulator and in real scenarios. With the same environment, RL-GSBridge exhibits behavior highly consistent with simulation tests during manipulation, whereas RL-sim without the GS model shows significant differences. Notably, blue lights in the camera view of Fig. \ref{fig:testbehaviour} is caused by the gripper indicator light. The image augmentation during policy training would mitigate this hue effect, ensuring the consistency of policy behavior. Meanwhile, RL-GSBridge sitll ensure the consistency of texture details between simulated and real-world images. 

\begin{figure}[t]
    \centering
    \includegraphics[width=0.85\linewidth]{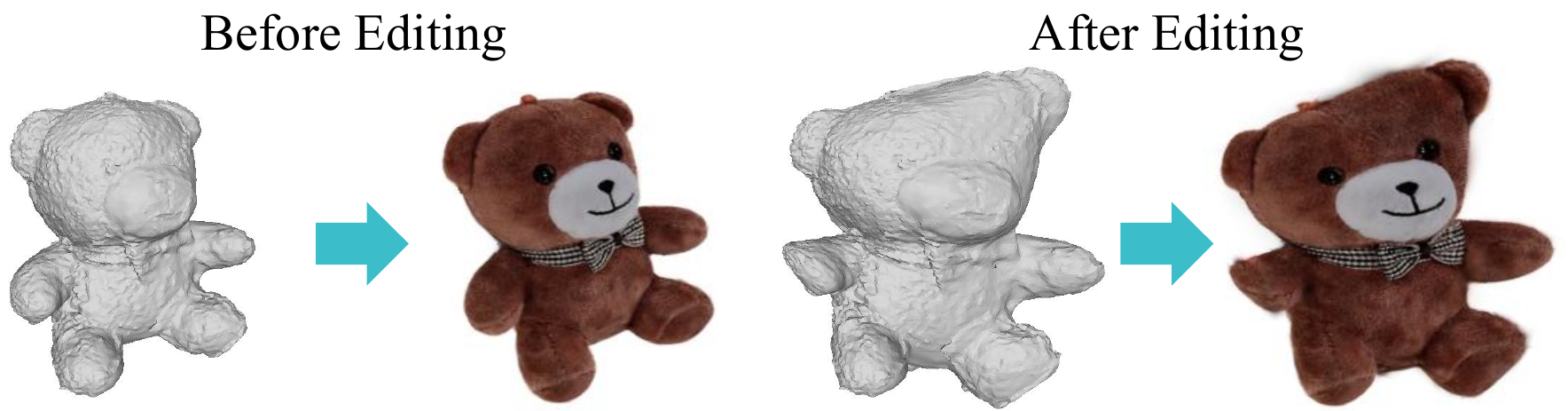}
    \caption{The editing capability on non-rigid objects of our soft binding constraint GS modeling method.}
    \label{fig:GSedit}
\end{figure}

\subsubsection{GS Rendering results of different Mesh Binding approach}\label{sec_soft}
In Table \ref{tab:GSresult}, we compare the performance of GaMeS \cite{waczynska2024games} and our soft mesh binding GS model under various scenarios and achieve the SOTA performance.
Furthermore, Fig. \ref{fig:GSresult} shows that our method obtains fewer artifacts and more detailed texture rendering.
As a supplement, in Fig. \ref{fig:GSedit}, our soft binding constraint GS modeling method achieves consistent results in editing a non-rigid toy bear. Unfortunately, due to the limitations in simulating soft objects in PyBullet, we do not demonstrate comprehensive experiments on deformable objects manipulation. However, this can be explored as a future direction.

\section{CONCLUSIONS}
We propose RL-GSBridge, a real-to-sim-to-real framework for robotic reinforcement learning. As an attempt to apply the recently successful radiance field reconstruction methods to construct a realistic robotic simulator, RL-GSBridge has shown promising sim-to-real success rates in desktop-level tasks. This motivates us to explore future directions, such as investigating the simulation of realistic lighting~\cite{gao2024relightable}, and integrating RL-GSBridge with advanced large-scale policy models~\cite{ma2024survey} and perception learning methods~\cite{mendonca2023structured, fang2023you, fang2025rethinking}.We hope RL-GSBridge will encourage more attempts to apply radiance field reconstruction methods in robotics.









\bibliographystyle{./IEEEtran} 
\bibliography{IEEEexample}

\end{document}